# Exploring the Unexplored: Understanding the Impact of Layer Adjustments on Image Classification

Haixia Liu[1], Tim Brailsford[1], James Goulding[2], Gavin Smith[2], and Larry Bull[1]

[1] University of the West of England, Bristol, UK
`haixia.liu@uwe.ac.uk`
[2] N/Lab, University of Nottingham, UK

Abstract. This paper investigates how adjustments to deep learning architectures impact model performance in image classification. Small-scale experiments generate initial insights although the trends observed are not consistent with the entire dataset. Filtering operations in the image processing pipeline are crucial, with image filtering before pre-processing yielding better results. The choice and order of layers as well as filter placement significantly impact model performance. This study provides valuable insights into optimizing deep learning models, with potential avenues for future research including collaborative platforms.

Keywords: Explainabe AI · Layer selection · Layter ordering · Filter Placement

# 1 Introduction

Deep learning applications have gained significant importance in recent years, with widespread use in areas ranging from Generative Pre-trained Transformers (GPT) to Residual Networks (ResNet). However, these algorithms are often viewed as "black boxes", making it difficult for individuals, especially those without expertise in AI techniques and terminology, to understand the reasoning behind their results. This has spurred growing interest in Explainable AI (XAI), an approach that seeks to clarify deep networks' processes and representations by frequently integrating both transparent and black box models.

Modifications to deep network architectures often involve adjusting individual layer specifics or reordering layer combinations. For instance, in image classification, deep learning networks typically consist of Batch Normalization (BN) layers, Convolutional Layers (Conv), Pooling Layers (PL), Dropout Layers (DropL), Fully Connected Layers (FCL), and Activation Layers (AL). Altering the details of these layers or changing their combinations can enhance model performance. XAI researchers work to better understand how these modifications enable the model to arrive at its predictions.

Training new models from scratch and exploring potential architectures can be extremely time-consuming, leading AI practitioners to seek shortcuts. They often employ standard models developed for one scenario in entirely different contexts. This pragmatic approach to network training is often used, and it has been demonstrated that fine-tuned models trained on ImageNet can be useful in other domains [23]. Although this work also showed that transferability diminishes as the distance between the base task and the target task increases.

In this paper, we demonstrate that even minor adjustments to existing models can substantially impact their performance. We employ small-scale experiments to generate initial 'inspirations' which can then be used to validate larger datasets and refine existing algorithms. Furthermore, we examine the effects of dataset characteristic differences on model performance. By focusing on the specific changes that lead to improved outcomes for various data collections, we are aiming to optimise image classification and gain a better insight into how models function with different datasets.

## 2 Related Work

In the influential DARPA XAI program ([11,12]), AI has been categorized into three waves: descriptive (i.e., symbolic reasoning); predictive (i.e., statistical learning); and explanatory (i.e., contextual adaptation). An example they give is an algorithm to recognize images of cats. The predictive outcome generates the probability that a given image is a cat, but an explainable approach focuses on an understandable human interface (i.e., it has fur, whiskers, and is of the cat class, therefore it is a cat). An important goal of XAI is to map models onto real-world considerations by the most effective choice of features. One useful aspect of this is to consider how much performance is lost if a given feature is not available. Leave One Feature Out (LOFO) is an algorithm that estimates the importance of a feature by iteratively removing each one from the set, and then evaluating the performance of the model, as used by [17]. This illustrates the importance of exploring datasets as well as choosing the right network architecture for the job, as discussed by [8].

In recent years, extensive work has been done on techniques for automating the optimization of network architectures [7]. Even partial automation of this process helps make deep learning more accessible to both researchers and practitioners. However, much of this work has focused on searching for the model with the best performance, rather than explaining why that performance is good especially where there is limited investigation of the datasets. A lot of work has been done on understanding the role of individual neurons and layers in deep neural networks (e.g. [2]), where they have been found to detect meaningful abstractions from images. The role of Activation Layers has been well explored and explained, particularly in ReLu (e.g.[20][3]), so it is not the focus of our current work. In addition, various aspects of dropout have been studied. Baldi and Sadowski (2013) developed a general formalism for analysing dropout and demonstrated that it performs stochastic gradient descent on a regularized error function [1]. Hahn and Choi (2020) have also suggested that dropout is effective because it generates more gradient information flow through the layers, pushing networks towards the saturation areas of nonlinear activation functions [13]. Bull and Liu (2022) have described the improved performance provided by dropout differently, in that it avoids local optima through learning in reduced dimensions [6].

Batch Normalization (BN) is a technique that normalizes activations in networks to improve training efficiency and stability, significantly speeding up training time, increasing learning rates, and improving model accuracy [16]. Ever since the introduction of BN, researchers have been trying to understand the mechanisms involved. Studies of the technique have found that the performance benefits are not solely due to internal covariate shift, but rather are due to the re-parameterization of the optimization problem, making it more stable and smoother [21,4].

Various studies have investigated the impact of combining different types of layers. Paradoxically, two of the most powerful techniques in deep learning, dropout and BN, usually have worse performances when combined than when used separately [19]. However, they can be complementary in some models, such as Wide ResNet (WRN). This reduced performance is attributed to "variance shift" resulting from inconsistent variances in the training and testing phases. The large feature dimension in WRN helps mitigate this issue. In another study, Gholamalinezhad and Khosravi (2020) compared various pooling techniques used with convolutional networks [9]. These have the potential to reduce spatial dimensions, decrease computational cost, and control overfitting while extracting useful information from input data. Jung et al. (2019) have found that it can be helpful for convolutional networks to restructure the BN layers by splitting and recombining them, and that non-convolutional layers play an important role in accelerating processing [18]. He et al.(2019) have studied various techniques for training deep convolutional networks, demonstrating consistent accuracy improvements across multiple architectures and boosting transfer learning performance in object detection and semantic segmentation[15]. However, all of these studies were done on limited datasets, and so it is hard to generalize rules. There is still plenty of scope for exploration, and a need to study the impacts of combining layers using different datasets.

Apart from the reordering of layers, adding or cutting them from a baseline model also has the potential to be useful. For example, Yang et al.[22] used a few lightweight adapters, which were added to the existing image model, and they found that these adapters can help achieve comparable or even better performance.

## 3 Scope and Contribution

In this study, we have focused entirely on image classification. Similar issues face other problems, such as natural language processing (NLP) or U-Net, but the results may well differ, and we are making no claims about generalisation. The motivation behind this work arises from the observation that powerful models are almost always developed incrementally over time, and the process of discovering these models tends to be expensive. The fundamental elements of these models are mathematical concepts, algorithms, and recombining and reordering these components can lead to new discoveries. It is, though, important to recognise that incorporating novel components can often have more significant impact.

For tasks, such as CIFAR-10 classification, deep learning solutions like ResNet-50 and ResNet-101 usually demonstrate superior performance compared to ResNet-18 due to their increased depth and capacity to learn complex features. An important issue to consider, when designing architectures, is which specific aspects of deep learning contribute to this improvement? For example, it might be in the organisation of layers and modification of filter sizes. With these questions in mind, we conducted experiments using a progressive learning and exploration approach. Our aim is to uncover insights from small-scale experiments, and ultimately either reinforce or challenge these findings by replicating them on a larger scale. Because conducting such studies on large datasets often requires formidable computational power, we are proposing a general framework that will enable AI researchers to explore these mechanisms with small datasets in order to make informed decisions as to whether more thorough (and costly) studies are worth doing.

The recent success of transformers in AI emphasizes the importance of exploring diverse component combinations in model architectures. Experimentation can lead to better understanding, improved performance, better generalization, new insights, adaptability, and innovation. We have clearly demonstrated a need for a framework that will facilitate this by providing a standardised sequence of operations for such comparisons, and we have outlined what this framework needs to consist of. The framework proposed is inspired by Brunner's discovery learning principles [5], which are based on constructivist theories of learning. The framework aims to enhance critical thinking and creative problem-solving, meaningfulness, exploration, and collaborative feedback. Specifically, the framework involves selecting an architecture based on prior knowledge, exploring various architectural permutations, and then analyzing and comparing the results at each stage.

## 4 Methodology

The experiments were implemented in Python using Keras. The data sets used were: CIFAR-10, Fashion MNIST, MNIST and MedMNIST. We started by experimenting using small datasets, observing, and comparing the results generated from each model. We considered whether these runs were consistent or interesting and we carried out the experiments on larger datasets to compare the outcomes from larger datasets with small datasets.

4.1 Operations

Data was subject to the following operations: image processing operations; layer operations and local operations. The image processing operations we considered were: resizing (upsampling); filtering (Sharpening or blurring) and reorder them. Image augmentation is not the focus of this study, thus, none of the experiments used image augmentation.

*Layer Operations*

– Plug-and-Play (PaP). These were used to evaluate the impact of single layers on different datasets
– Leave One Layer Out (LOLO). This was used to evaluate the impact of a single layer on different datasets and conversely on for comparison to PaP.
– Select-and-Reorder(SaRe). This was used to investigate the impact of two or more layers' placement, on different datasets.
– Special investigation on BN and DropL.

*Local Operations*

– Filter placement. This was used to assess the influence of the arrangement of filter size. Specifically, the change in filter size is being tested by altering its size from an increasing sequence Res64to512(64, 128, 256, 512) to a decreasing sequence Res512to64(512, 256, 128, 64). This allows us to examine the effect of reducing or increasing the filter size. In addition, Res64(64, 64, 64, 64) and Res512(512, 512, 512, 512) are also being compared.

4.2 Exploration

We explore the impact of the three base models and operations on the different datasets. Base0 is a fully connected simple neural network (FCL). BaseSeq model: Conv-BN-PL-DropL-FCL-AL. BaseRes18 is the ResNet18[14] model.

Plug and Play (PaP) is used with Base0 to evaluate the impact of a single layer. Using the BaseSeq model we performed LOLO to evaluate the impact of a single layer, and also SaRe to investigate the impact of the ordering of pairs of layers. For BaseRes18, we explored the filter placement.

## 4.3 Datasets

We constructed the following sub-datasets: From each base dataset (MNIS, FashionMNIST, and CIFAR-10), we randomly sampled 128/600 and used full samples from hardclass and easyclass pairs. To identify hard and easy classes, we analyzed the confusion matrix from published studies, for example [10].

Table 1 displays the resulting subset characteristics, and sampleSize denotes the sum of training and validation samples. We applied *stratify=y* when splitting the dataset to maintain the same class ratio in both the training and validation subsets. Note that we converted FashionMNIST(FMNIST) and MNIST images to three channels for ResNet-related experiments.

| Subset name | Class names (sample size) | Metadata |
|---|---|---|
| Cifar10Hard2Cls128/600/All | dog(128/600/12000) vs cat(128/600/12000) | 32by32, 3channel |
| Cifar10Easy2Cls128/600/All | car(128/600/12000) vs deer(128/600/12000) | 32by32, 3channel |
| MNISTHard2Cls128/600/All | 4(128/600/6824) vs 9(128/600/6958) | 28by28, grayscale |
| MNISTEasy2Cls128/600/All | 1(128/600/7877) vs 5(128/600/6313) | 28by28, grayscale |
| FMNISTHard2Cls128/600/All | T-shirt/top(128/600/7000) vs Shirt(128/600/7000) | 28by28, grayscale |
| FMNISTEasy2Cls128/600/All | Ankle boot(128/600/7000) vs Bag(128/600/7000) | 28by28, grayscale |

Table 1: Small datasets for exploration, using different subjects (classes) and sample sizes.

After completing these small scale experiments we used all samples from CIFAR-10, MNIST, and FashionMNIST. Finally we tested what we had discovered by repeating the process using MedMNIST.

## 4.4 Experiment Settings

The parameters used across all experiments were as follows:

*testsize=0.25, randstate=42, batchsize=16, nepochs=100,
validationsplit=0.25, dropoutrate=0.25, verbosity=1*

Please note that all reported accuracy scores are based on the test sets and were obtained using EarlyStopping with a patience value of 10.

## 5 Results

We initially explored the potential for valuable insights from small-scale datasets, specifically those containing 128 or 600 samples from selected pairs of classes. Figures 1a and 1b depict the accuracy of various models using different sub-datasets. The modified models showed some impact on CIFAR-10 and Fashion MNIST samples but had minimal impact on MNIST. As a result, we focused our investigation on CIFAR-10 and Fashion MNIST, rather than MNIST.

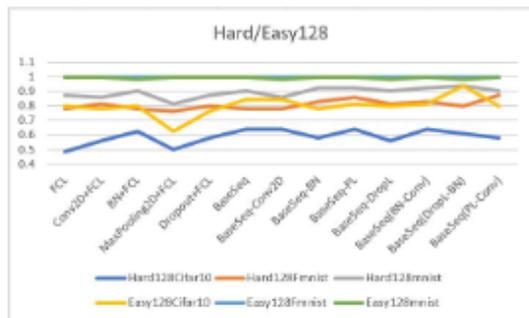
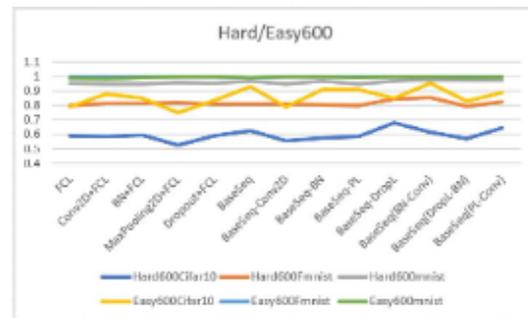

(a) Accuracy of different models on Hard128/Easy128 sub-datasets.

(b) Accuracy of different models on Hard600/Easy600 sub-datasets.

Fig. 1: Results of small scale experiments using 128 and 600 sample sizes.

To investigate the consistency of fluctuations, we hypothesized that using a sample size of 128 would not produce consistent results since each run would involve different samples. Figures 2a, 2b and 2c support this observation. With larger sample sizes, Figures 3a and 3b show the accuracy of different models (Base0 and BaseSeq variations) using the subsets as well as the entire dataset. Figures 4a and 4b display the accuracy of different models (BaseRes18 variations) on subsets and the full dataset. Our goal in analyzing these results was to determine if specific subsets could reveal trends similar to those observed in the entire dataset. We found that the HardAll subset better aligned with the full dataset. Additionally, decreasing the filter size from 512 to

64 and using a straight 512 filter size placement led to improved CIFAR-10 results.

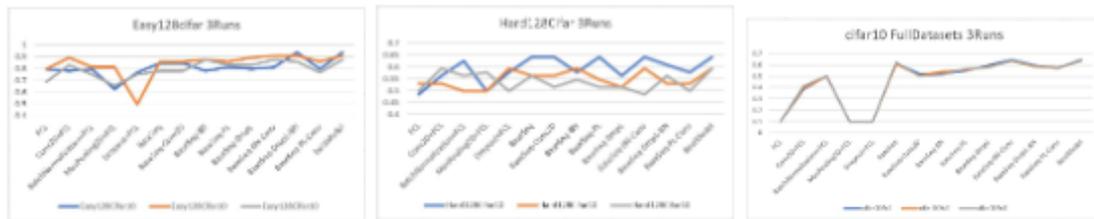

(a) 3 runs of CIFAR-10 (Easy128).  (b) 3 runs of CIFAR-10 (Hard128).  (c) 3 runs of CIFAR-10 (entire).

Fig. 2: Results of the fluctuation consistency experiment (3 runs).

Figure 4a and 4b show the accuracy from different models (BaseRes18 variations) on the subsets and the entire set. Our goal in analyzing these results was to determine if specific subsets could reveal trends similar to those observed in the entire dataset. We found that the HardAll subset is best aligned with the full dataset. Additionally, decreasing the filter size from 512 to 64 and using a straight 512 filter size placement led to improved CIFAR-10 performance.

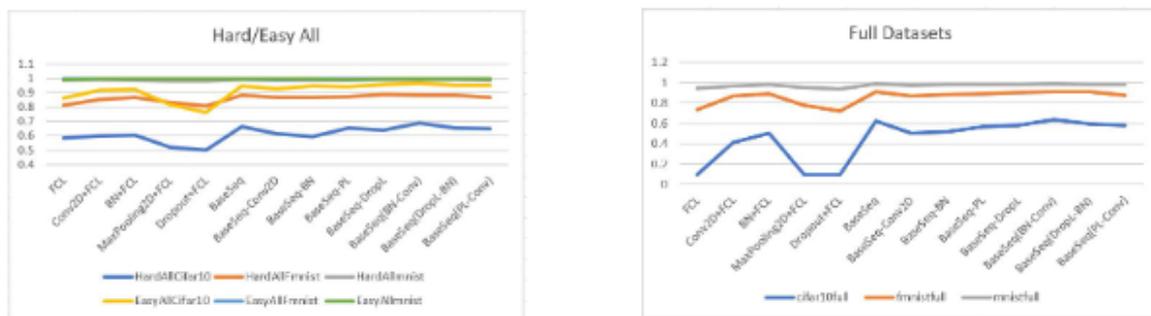

(a) Accuracy of models using HardAll/EasyAll sub datasets.  (b) Accuracy of models using entire sets.

Fig. 3: Results of different models based on Base0 and BaseSeq using larger datasets (HardAll/EasyAll vs entire set).

Figures 5a and 5b illustrate the variations in accuracy when comparing the sequence of batch normalization and dropout layers, using Base0, which includes a single fully connected layer (FCL). Similar trends can be observed between the HardAll/EasyAll and entire set categories, with the HardAll subset offering a slight advantage in this study. However, the Easy600 subset, did not accurately reflect the impact of the models on the complete dataset.

These findings suggest that relying solely on small-scale experiment results may not be sufficient; instead, we should conduct binary analyses of small set outcomes. If the results change when using modified models, it will be important to conduct further experiments using larger datasets.

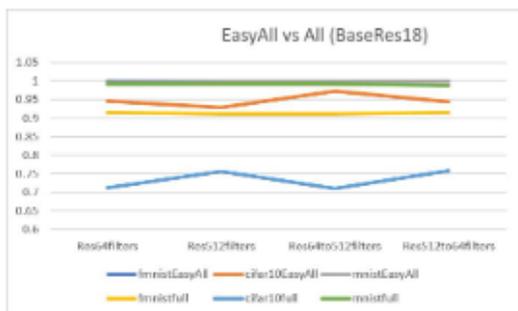
(a) Accuracy of models using EasyAll and entire sets.

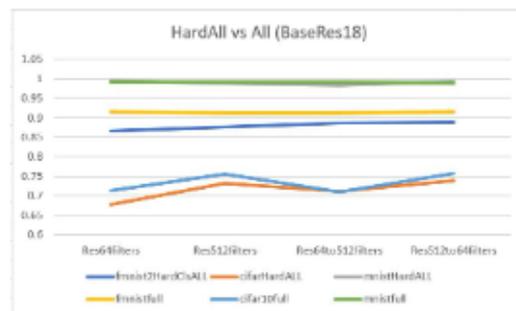
(b) Accuracy of models using HardAll and entire sets.

Fig. 4: Results of filter placement based on BaseRes18 modifications.

The results of image processing effects and other findings are presented in Table 2.These results indicate that applying filters prior to pre-processing leads to improved performance. The preprocessing utilised the preprocess input function from TensorFlow[5]. The filters explored in this study include sharpening and blurring, and also the UpSampling2D layer from Keras was used.

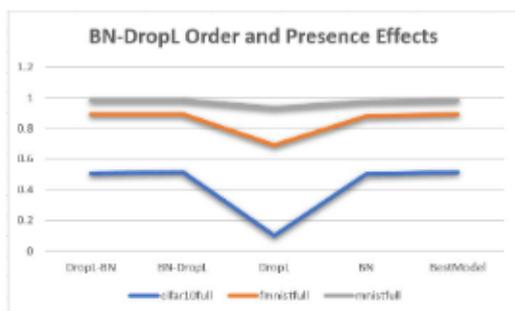
(a) Ordering effects and presences of layers (BN and dropL).

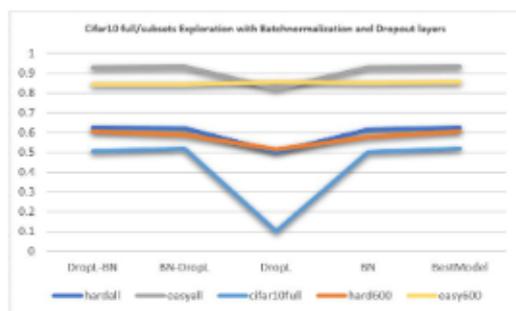
(b) Comparing the impact of sample size and Hard/Easy categorization using CIFAR-10.

Fig. 5: Effects of the ordering of layers (BN and DropL).

| Presence and ordering or operations (image processing) | Data |
|---|---|
| Filter-Preprocessing > Preprocessing-Filter | Cifar10HardFull |
| With UpSampling2D > Without UpSampling2D | Cifar10HardFull |
| Conv-PL-UpSamp,repeated 3 and 7 times > Conv-PL-UpSamp-DropL,repeated 3 and 7 times | Cifar10Full |

Table 2: Effects of image processing and model architecture on performance.

We explored the following operations on the subsets of the MedMNIST datasets: Conv2d+FCL, BN+FCL, BaseSeq, BaseSeq(BN-Conv), Res64to512(original ResNet18), Res512, Res512to64 on MedMNIST datasets. Entropy was calculated for each dataset. The results are shown in Table 3.

| Data | BestModel Among Selected Base0 and BaseSeq Variations | BestModel Among BaseRes18 Variations | Entropy |
|---|---|---|---|
| pathmnist | BaseSeq(BN-Conv) | Res512to64 | -8.41 |
| dermamnist | Conv2D+FCL | Res512 | -8.64 |
| octmnist | BaseSeq(BN-Conv) | Res512to64 | -1.26 |
| pneumoniamnist | BaseSeq(BN-Conv) | Res512 | 0.32 |
| breastmnist | Conv2D+FCL | Res512 | 0.33 |
| bloodmnist | Conv2D+FCL | Res64to512 | -6.99 |
| tissuemnist | BaseSeq(BN-Conv) | Res512to64 | -5.93 |
| organamnist | BaseSeq(BN-Conv) | Res512 | -14.90 |
| organcmnist | BaseSeq(BN-Conv) | Res512to64 | -14.05 |
| organsmnist | BaseSeq(BN-Conv) | Res64to512 | -14.51 |

Table 3: Best performing models using MedMNIST.

In Figures 3b, the results of BaseSeq and BaseSeq(BN-Conv) are close. Similarly, the results of Res512 and Res512to64 are close to each other in Figure 4b. Table 4 shows the detailed accuracy from each model.

# 6 Discussion

In this study, we analysed the consistency of findings between different dataset sizes, comparing the results from smaller subsets to those obtained from the full datasets. Consistent findings across dataset sizes would enable researchers to test novel methods on smaller datasets, providing a valuable shortcut when computational resources are limited. Our analysis revealed that trends derived from small data subsets (128, 600 samples from two selected classes) do not consistently align with those observed when using the entire dataset. However, the benefit of exploring smaller datasets is that if model adjustments impact performance on smaller datasets, it is likely to be worth conducting further experiments on larger datasets to validate those findings.

| Model(FirstRun/SecondRun/ThirdRun) | Entire Dataset |
|---|---|
| BaseSeq(Conv-BN)(0.622/0.616/0.607) | CIFAR-10 |
| BaseSeq(BN-Conv)(**0.637/0.643/0.632**) | CIFAR-10 |
| Res64to512(0.71/0.721/0.702) | CIFAR-10 |
| Res512to64(**0.758**/0.732/0.748) | CIFAR-10 |
| Res512(0.756/**0.769/0.757**) | CIFAR-10 |
| Res64(0.901/0.912/0.908) | FMNIST |
| Res64to512(0.909/0.902/0.905) | FMNIST |
| Res512to64(0.909/0.914/0.916) | FMNIST |
| Res512(**0.916/0.918/0.918**) | FMNIST |
| Res64(0.991/0.991/0.991) | MNIST |
| Res64to512(**0.992/0.992/0.992**) | MNIST |
| Res512to64(0.99/0.988/0.99) | MNIST |
| Res512(0.991/**0.992**/0.989) | MNIST |

Table 4: Aggregate outcomes from 3 experiments using the full datasets. CIFAR-10 sees improved performance when the Conv and BN layers switch places. CIFAR-10 and FashionMNIST are improved by Res512 or Res512to64 rather than ResNet18's default filter size placement of Res64to512. MNIST has a slight inclination towards Res64to512. The best results are highlighted in bold.

The ordering of filtering and pre-processing operations in the image processing pipeline is important. Performing image filtering before image pre-processing yielded better results than the other way around, and incorporating upsampling into the processing pipeline improved model performance. Images are converted from RGB to BGR, and then each color channel is zero centered without scaling. We noticed that the pre-processed images became much cooler. Color temperature, an important concept in photography, is linked to light's spectral composition and has a major impact on how humans perceive images. It is possible that this might affect algorithmic perception. Potentially, this could explain the impact of pre-processing on classification. Further research is needed in this area, but if it is found to be true, then it would be important to explainable AI (XAI), because it could map part of the model to the physics of light.

Not only is the choice of layers important, but so is their ordering. When adding one layer to the FCL, BN provides the best performance, but when removing one layer from the BaseSeq, removing Conv resulted in the worst outcome. Upon examining combinations of BN and DropL, it's different from the outcomes by the work [19], which used the CIFAR-10/100 dataset but different experiment settings as we did. Furthermore, the BN-Conv order was superior to the Conv-BN order for most datasets. Filter size placements of Res512to64 and Res512 outperformed the ResNet18 default placement (64, 128, 256, 512) in most, but not all, cases.

The main limitations of this study are a reliance on a limited dataset and repetitions, making statistical analysis difficult. However, in the future we hope to use more detailed dataset characteristics to help elucidate the observed trends and findings. It should be possible to use mathematical approaches to derive an understandable prototype and enhance model explainability through a comprehensive exploration framework. Moreover, a collaborative platform could enable researchers and engineers worldwide to work together, regardless of their computational resources, to explore uncharted areas and collaboratively tackle challenging problems in the field of deep learning.